\documentclass[letterpaper, 10 pt, conference]{ieeeconf}  % Comment this line out if you need a4paper

\IEEEoverridecommandlockouts                              % This command is only needed if 
\usepackage{booktabs}
\usepackage{multirow}
\usepackage{adjustbox}
\usepackage{caption}
\usepackage{amsmath}
\usepackage{booktabs}
\usepackage{siunitx}
\usepackage{graphicx}
\usepackage{amsfonts}
\usepackage{subcaption}
\usepackage{hyperref}
\usepackage{cite}
\title{\LARGE \bf
Diffusion-Based Low-Light Image Enhancement\\with Color and Luminance Priors
}

\author{Xuanshuo Fu, Lei Kang$^{\dag}$ and Javier Vazquez-Corral% <-this % stops a space
\thanks{This work has been partially supported by the predoctoral program AGAUR-FI ajuts (2025 FI-2 00470) Joan Oró, which are backed by the Secretariat of Universities and Research of the Department of Research and Universities of the Generalitat of Catalonia, as well as the European Social Plus Fund; the Beatriu de Pinós del Departament de Recerca i Universitats de la Generalitat de Catalunya (2022 BP 00256); the Grant PID2024-162555OB-I00 funded by MCIN/AEI/10.13039/501100011033, ERDF ``A way of making Europe'' and by the Generalitat de Catalunya CERCA Program. JVC also acknowledges the 2025 Leonardo Grant for Scientific Research and Cultural Creation from the BBVA Foundation. The BBVA Foundation accepts no responsibility for the opinions, statements and contents included in the project and/or the results thereof, which are entirely the responsibility of the authors.}% <-this % stops a space
\thanks{Computer Vision Center, Cerdanyola del Vallès, Spain and Universitat Autònoma de Barcelona, Cerdanyola del Vallès, Spain}
\thanks{\tt \footnotesize \{xuanshuo, lkang, javier.vazquez\}@cvc.uab.cat}%
\thanks{ \dag denotes Corresponding Author.}%
}

\begin{document}

\maketitle
\thispagestyle{empty}
\pagestyle{empty}

%%%%%%%%%%%%%%%%%%%%%%%%%%%%%%%%%%%%%%%%%%%%%%%%%%%%%%%%%%%%%%%%%%%%%%%%%%%%%%%%
\begin{abstract}
Low-light images often suffer from low contrast, noise, and color distortion, degrading visual quality and impairing downstream vision tasks. We propose a novel conditional diffusion framework for low-light image enhancement that incorporates a Structured Control Embedding Module (SCEM). SCEM decomposes a low-light image into four informative components including illumination, illumination-invariant features, shadow priors, and color-invariant cues. These components serve as control signals that condition a U-Net–based diffusion model trained with a simplified noise-prediction loss. Thus, the proposed SCEM equipped Diffusion method enforces structured enhancement guided by physical priors. In experiments, our model is trained only on the LOLv1 dataset and evaluated without fine-tuning on LOLv2-real, LSRW, DICM, MEF, and LIME. The method achieves state-of-the-art performance in quantitative and perceptual metrics, demonstrating strong generalization across benchmarks. \href{https://casted.github.io/scem/}{https://casted.github.io/scem/}.

\end{abstract}

%%%%%%%%%%%%%%%%%%%%%%%%%%%%%%%%%%%%%%%%%%%%%%%%%%%%%%%%%%%%%%%%%%%%%%%%%%%%%%%%
\section{INTRODUCTION}

% \begin{figure}
%   \centering
%   \includegraphics[width=0.9\linewidth]{Table1Radarchart.png}
%   \caption{}
%   \Description{}
%   \label{fig:first}
% \end{figure}

Low-light image enhancement (LLIE) aims to recover a clean, perceptually faithful normal-light image from severely underexposed, noisy input. Reliable LLIE is critical for nighttime photography, mobile vision, surveillance, and autonomous systems in which downstream perception degrades when contrast collapses, color shifts, and sensor noise dominate~\cite{xu2023low}. Real scenes further complicate the task: illumination varies spatially, shadows interact with mixed light sources, and color responses drift nonlinearly across the image~\cite{zhou2025low}. Effective methods must therefore brighten selectively while preserving structure, texture, and plausible color.

Classical approaches manipulate pixel statistics or hand-crafted illumination models. Histogram equalization expands dynamic range but readily amplifies noise and creates unnatural luminance~\cite{abdullah2007dynamic}. Retinex-style decompositions separate illumination from reflectance and can restore detail, yet they are sensitive to parameterization and may introduce halos or color artifacts~\cite{li2011image}. Point-wise nonlinear curves (gamma, sigmoid) lack spatial awareness and struggle under complex lighting.

\begin{figure}[t!]
    \centering
    \begin{subfigure}{0.48\linewidth}
        \centering
        \includegraphics[width=\linewidth]{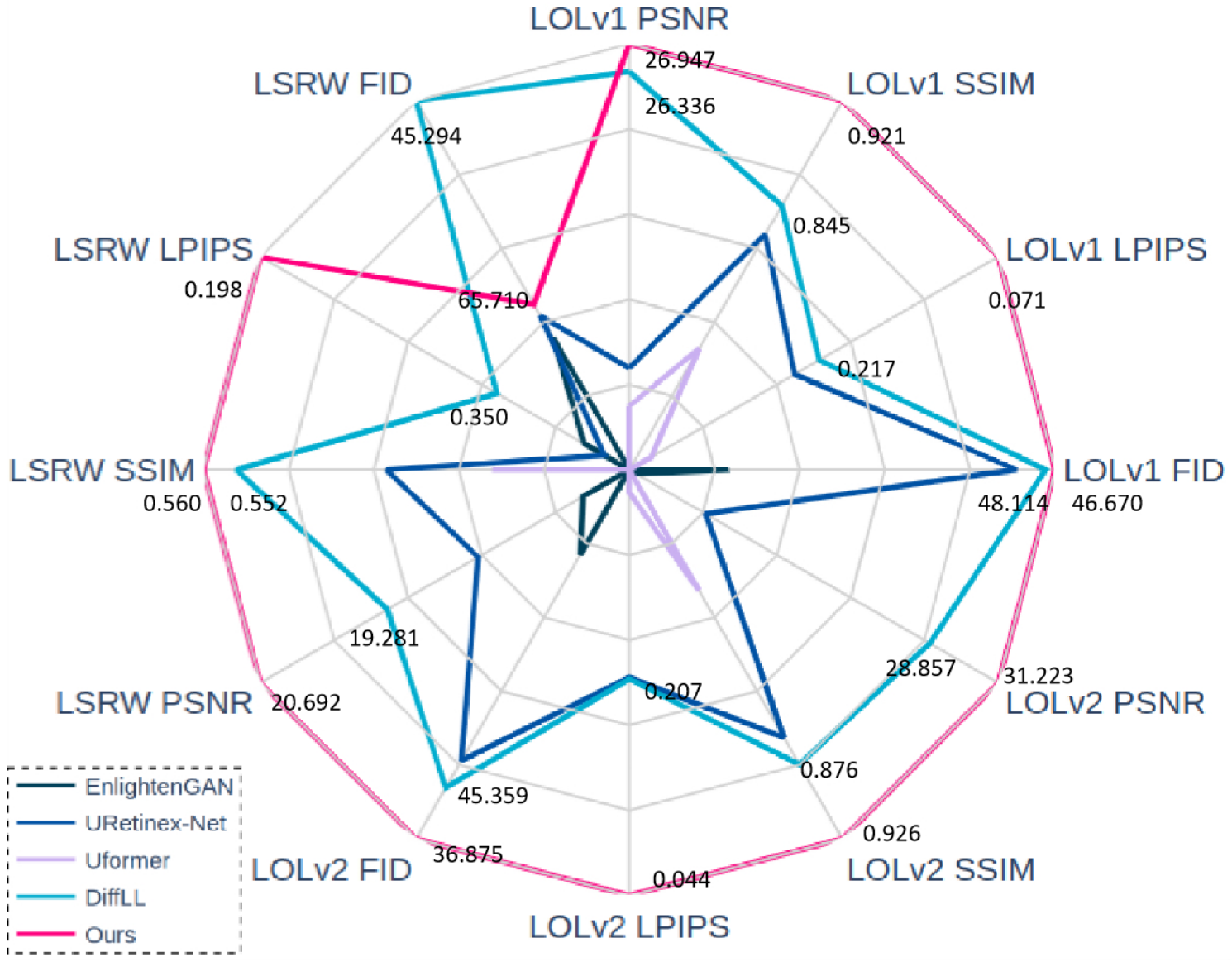}
        \caption{}
        \label{fig:subfig1}
    \end{subfigure}
    \hfill
    \begin{subfigure}{0.48\linewidth}
        \centering
        \includegraphics[width=\linewidth]{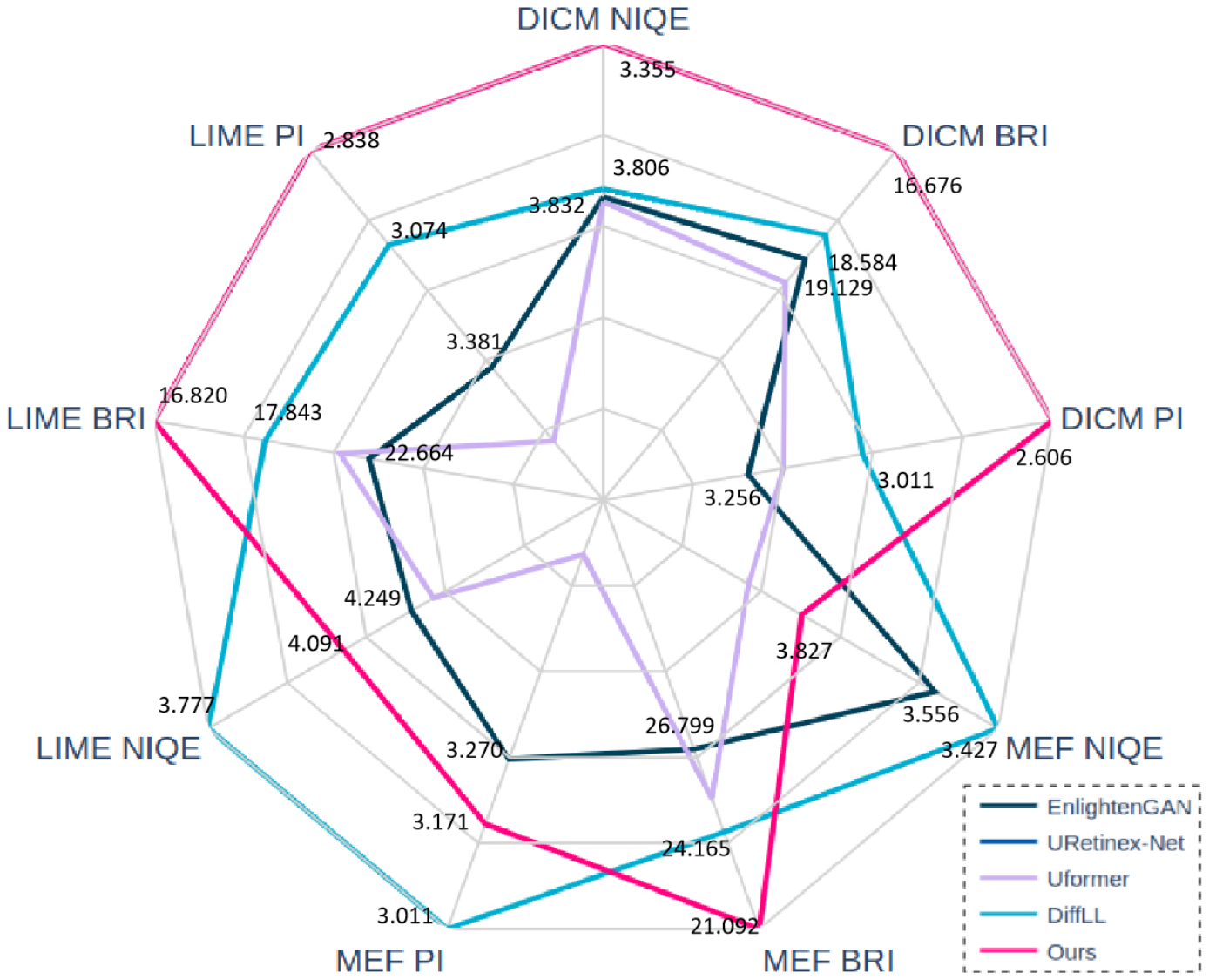}
        \caption{}
        \label{fig:subfig2}
    \end{subfigure}
    \caption{Quantitative comparisons with state-of-the-art methods. (a) presents numerical scores for PSNR, SSIM, LPIPS, and FID on 3 datasets: LOLv1, LOLv2-real, and LSRW, which contain groundtruth normal-light images. (b) presents numerical scores for NIQE, BRISQUE and PI on 3 datasets: DICM, MEF, and LIME, which contain only low-light images without groundtruth normal-light counterparts. Note that our proposed methods are trained only on the LOLv1 training set and are directly evaluated on the remaining datasets. To enable intuitive comparison in the radar plot, we normalized all metrics and inverted those where lower is better (e.g., LPIPS, FID), so that higher values consistently indicate better performance.}
   % \vspace{-2mm}
    \label{fig:radar}
\end{figure}

Learning-based LLIE substantially improves fidelity but often remains weakly grounded in image formation. CNN regressors (e.g., U-Net, ResNet variants) map low-light to normal-light images directly, treating enhancement as black-box translation that can overfit and hallucinate colors~\cite{wei2018deep,zamir2020learning}. GAN formulations encourage realism but inherit adversarial instability and may globally remap appearance~\cite{jiang2021enlightengan}. Diffusion models offer stronger generative coverage and training stability~\cite{zhou2025low}, yet vanilla conditional diffusion provides limited control over illumination consistency and color faithfulness when applied naively to LLIE.

We introduce a \textbf{Structured Control Embedding Module (SCEM)} that injects physically motivated, spatially aligned priors into a conditional diffusion backbone for LLIE. From each low-light input, SCEM computes four complementary guidance maps: (1) illumination to steer exposure balancing, (2) illumination-invariant features approximating reflectance for structure retention, (3) shadow priors to protect texture in dark and bright transitions, and (4) color-invariant cues that stabilize chromatic 
relationships. These maps are encoded and fused with U-Net features at every denoising step, making enhancement explicitly guided by scene lighting and color statistics rather than emergent from implicit regression. The design links Retinex-like reasoning with the expressive sampling power of diffusion, yielding more controllable and physically plausible results.

Trained solely on LOLv1, our SCEM-conditioned diffusion model generalizes without dataset-specific tuning to the public benchmarks LOLv2-real, LSRW, DICM, MEF, and LIME. It achieves state-of-the-art or highly competitive scores on the reference metrics PSNR and SSIM and on the perceptual metrics LPIPS, FID, NIQE, BRISQUE, and PI, yielding visibly sharper detail and more natural color reproduction.

Our contributions are threefold:
\begin{enumerate}
\item We propose SCEM, a structured control interface that embeds multi-channel illumination and appearance priors directly into a diffusion-based LLIE model, providing fine-grained, spatially aware guidance during denoising.
\item We operationalize a Retinex-informed decomposition jointly with shadow and color-invariance cues, enabling adaptive brightness boost while preserving texture and chromatic fidelity.
\item Extensive cross-dataset experiments show strong generalization from training on LOLv1 only and state-of-the-art performance on multiple LLIE benchmarks in both reference and no-reference metrics.
\end{enumerate}

\section{RELATED WORK}

\begin{figure*}
  \centering
  \includegraphics[width=0.87\linewidth]{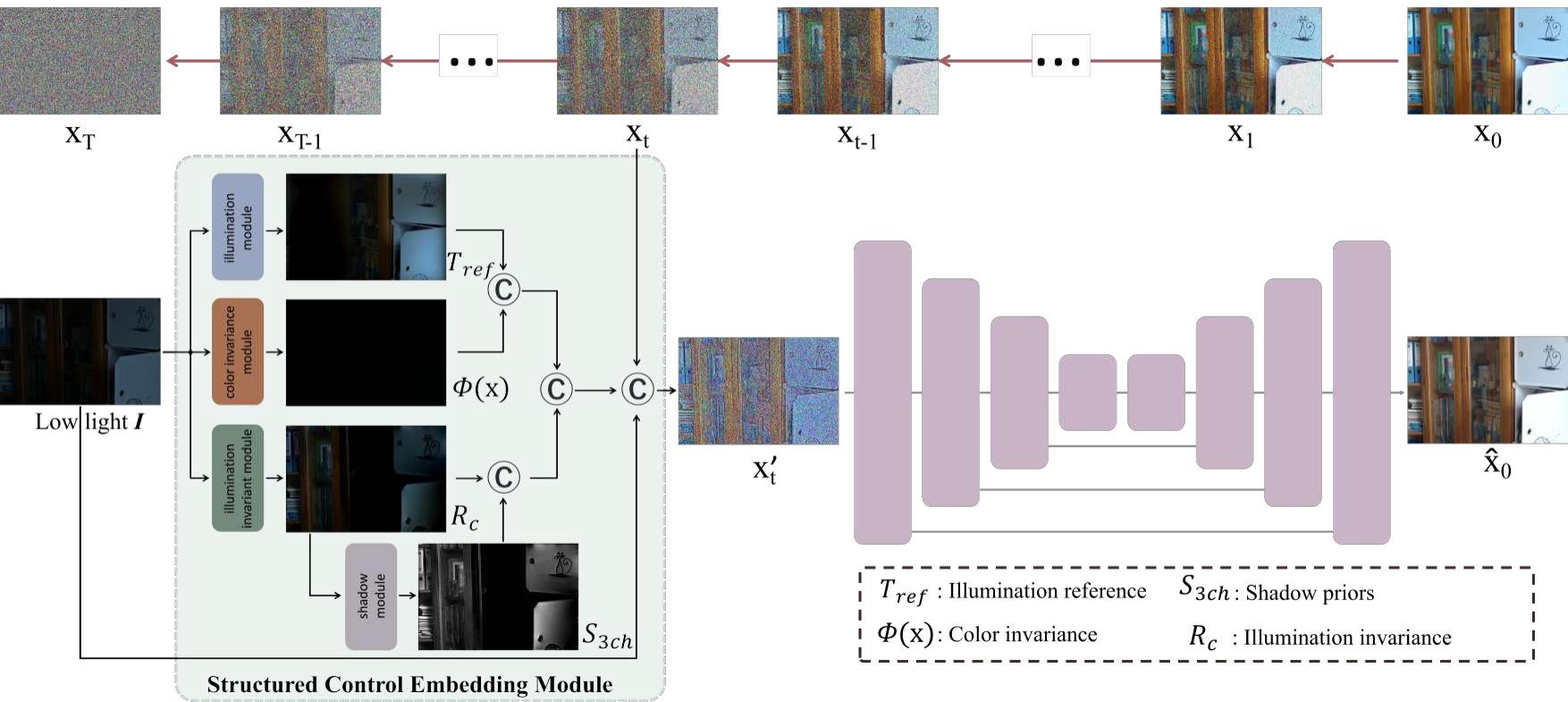}
  \caption{The proposed architecture of SCEM equiped diffusion model for low-light image enhancement. During training, we extract four types of features from the input low-light image $\mathcal{I}$: illumination $T_{ref}$, color invariance features $\Phi(x)$, illumination-invariant features $R_c$, and shadow priors $S_{3ch}$. These features, along with the original low-light image, are concatenated with the randomly chosen $t$-th noised image $X_t$ to form the input $X_t'$ for the denoising training process. The diffusion model then generates the enhanced image $\hat{X}_0$. During inference, Gaussian noise $X_T$ is concatenated with the same set of extracted features and the original low-light image $I$.}
  \vspace{-3mm}
  \label{fig:net}
\end{figure*}

% \textbf{Traditional Enhancement Methods:} Early low-light enhancement relied on image processing priors. Histogram equalization (HE)~\cite{jebadass2022low} and its variants (CLAHE, bi-histogram equalization, etc.) adjust pixel intensities to boost global contrast, but often over-enhance bright regions and amplify noise in dark regions. Retinex theory~\cite{land1971lightness} posits that an image can be decomposed into illumination (smooth component) and reflectance (texture/detail). Classical Retinex algorithms (SSR~\cite{land1971lightness}, MSR~\cite{jobson1997properties}, MSRCR~\cite{jobson1997multiscale}) combine multi-scale filtering to approximate this decomposition. These methods recover details from shadows but may introduce halo artifacts and color distortions. Follow-up techniques refined the filters (guided filters, bright-pass filters) and combined Retinex with gamma correction~\cite{jeon2024low} for better brightness stretching. Non-linear tone-mapping functions~\cite{yang2018low} such as gamma or logarithmic transforms have also been applied. While these approaches are computationally light and sometimes effective, they lack learned adaptation and cannot fully eliminate noise or complex color casts, motivating data-driven enhancements.

\textbf{Traditional Enhancement Methods:} Classical low-light enhancement relied on hand-crafted image processing. Histogram equalization (HE)~\cite{jebadass2022low} and derivatives rescale intensities to raise contrast but often blow out highlights and amplify noise in dark regions. Retinex theory models an image as illumination $\times$ reflectance~\cite{land1971lightness}. MSR/MSRCR~\cite{jobson1997properties}) approximate this separation via multi-scale filtering, recovering shadow detail yet frequently introducing halos and color shifts. Subsequent refinements improved the filtering and paired Retinex with gamma correction~\cite{jeon2024low} to better redistribute brightness. Simple non-linear tone mappings~\cite{yang2018low} provide additional contrast stretching. Despite being lightweight and sometimes effective, these rule-based schemes lack learned adaptability and struggle to suppress noise or correct complex color casts, motivating data-driven methods.

% \textbf{CNN-based Methods:} Deep learning approaches learn complex mappings for enhancement using large datasets. RetinexNet~\cite{wei2018deep} was among the first CNNs for low-light imaging: it learns a decomposition network and an enhancement network on paired low/normal-light images. The end-to-end pipeline yields visually pleasing results, but it requires supervised normal-light references and may not generalize beyond the training distribution. Unsupervised CNNs like Zero-DCE~\cite{guo2020zero} reformulate enhancement as estimating an image-specific tone-mapping curve. Zero-DCE’s DCE-Net is trained with non-reference losses (e.g., spatial consistency, exposure), eliminating the need for ground-truth images. Despite its flexibility, Zero-DCE can struggle under extreme lighting or introduce tonal artifacts if the learned curve is misestimated. More recently, multi-scale networks like MIRNet~\cite{zamir2020learning} incorporate parallel convolutional streams and attention to capture both fine details and high-level context. MIRNet’s multi-resolution residual blocks achieve state-of-the-art denoising and enhancement by learning enriched features that fuse global context with spatial detail. While powerful, these CNNs typically do not include explicit illumination modeling beyond what is learned implicitly, and they may overfit to specific datasets or fail to provide intuitive control over brightness.

\textbf{CNN-based Methods:} Deep networks learn data-driven enhancement mappings. RetinexNet~\cite{wei2018deep} jointly trains decomposition and enhancement subnets on paired low/normal-light data to produce pleasing results, but relies on supervised references and may not generalize. Zero-DCE~\cite{guo2020zero} removes paired data by learning an image-adaptive tone curve via non-reference losses (spatial consistency, exposure, etc.), yet can falter under extreme illumination and introduce tonal artifacts. Multi-scale/attention architectures such as MIRNet~\cite{zamir2020learning} fuse cross-resolution features to capture detail and context, yielding strong denoising and enhancement. Still, most CNNs only implicitly model illumination, risk dataset bias, and provide limited user control over brightness.

% \textbf{Transformer-based Methods:} Uformer~\cite{wang2022uformer} adopts a U-Net inspired encoder-decoder Transformer backbone with locally-enhanced window self-attention and a learnable multi-scale bias modulation strategy, allowing efficient capture of local and global features to recover fine details in dark images. Likewise, Restormer~\cite{zamir2022restormer} introduces a channel-wise attention design via multi-Dconv head transposed attention and gated depthwise feed-forward networks, enabling high-resolution low-light restoration with significantly reduced complexity. LLFormer~\cite{wang2023ultra} leverages axial multi-head self-attention and cross-layer feature fusion blocks to handle ultra-high-definition low-light images, cutting down the self-attention complexity while improving enhancement fidelity. While transformer-based methods offer strong context modeling and competitive results, they often require careful architectural tuning and large-scale training data, and they lack the generative flexibility and controllable sampling mechanisms offered by recent generative approaches.

\textbf{Transformer-based Methods:} Uformer~\cite{wang2022uformer} extends a U-Net encoder–decoder with locally enhanced window self-attention and learnable multi-scale bias modulation to efficiently capture local and global context for low-light detail recovery. Restormer~\cite{zamir2022restormer} employs channel-focused multi-Dconv head transposed attention and gated depthwise feed-forward blocks to restore high-resolution low-light images with reduced complexity. LLFormer~\cite{wang2023ultra} scales to ultra-high-definition inputs via axis-based self-attention and cross-layer feature fusion that lower attention cost while preserving fidelity. Despite strong context modeling and competitive results, Transformer models often demand careful tuning and large training corpora, and they still lack the generative flexibility and controllable sampling offered by recent generative approaches.

% \textbf{GAN-based Methods:} Generative adversarial models have been applied to low-light enhancement, especially in unpaired settings. EnlightenGAN~\cite{jiang2021enlightengan} trains an unsupervised GAN where the generator brightens the image and the discriminator enforces realism. Its contributions include a global-local discriminator and self-regularized perceptual loss that guide enhancement without paired data. The result is a flexible model that adapts to various domains, although GAN training can be unstable and may produce artifacts (e.g. color shifts or oversmoothing). ReLLIE~\cite{zhang2021rellie} uses reinforcement learning to sequentially estimate pixel-wise enhancement curves. Here, a policy network decides how much to brighten each pixel step-by-step, driven by carefully crafted no-reference reward functions. This approach handles diverse lighting conditions and can be “paused” when visually satisfied, offering user-specific control. However, ReLLIE and similar methods do not explicitly decompose image layers; the learned policy implicitly entangles illumination and reflectance, which can limit interpretability.

\textbf{GAN-based Methods:} GANs tackle low-light enhancement, especially when paired supervision is scarce. EnlightenGAN~\cite{jiang2021enlightengan} trains an unpaired generator to brighten inputs while a global--local discriminator and self-regularized perceptual loss promote realistic results across domains, though adversarial training can be unstable and introduce color or smoothing artifacts. ReLLIE~\cite{zhang2021rellie} casts enhancement as sequential pixel-wise curve adjustment via a reinforcement-learned policy driven by no-reference rewards; users can effectively halt when visually satisfied, but the policy lacks explicit illumination/reflectance separation, limiting interpretability.

\textbf{Diffusion-based Methods:} Denoising diffusion probabilistic models (DDPM)~\cite{ho2020denoising} and their accelerated variants like denoising diffusion implicit models (DDIM)~\cite{song2020denoising} learn to reverse a gradual noise process. Conditioning diffusion models on inputs enables powerful image-to-image translation. Palette~\cite{saharia2022palette} demonstrates that a unified diffusion framework can outperform task-specific GANs across colorization, inpainting, and super-resolution. In low-light enhancement, CLE Diffusion~\cite{yin2023cle} introduces user-controllable enhancement via illumination and semantic conditioning, while DiffLL~\cite{jiang2023low} improves efficiency and texture fidelity using wavelet-domain diffusion and high-frequency refinement. GDP~\cite{fei2023generative} treats enhancement as blind restoration, optimizing degradation parameters during sampling. Diff-Retinex~\cite{yi2023diff} integrates Retinex-based decomposition with conditional diffusion to jointly address illumination, noise, and structural loss, enabling detailed content recovery and physically interpretable enhancement.

%%%%%%%%%%%%%%%%%%%%%%%%%%%%%%%%%%%%%%%%%%%%%%%%%%%%%%%%%%%%%%%%%%%%%%%%%%%%%%%%%%%%%%%%%%%%%%%%%%%%%%%%%%%%%%%%%%%%%%%%%%%%%%%%%%%%%%%%%%%%%%%
\section{METHODOLOGY}

% Our proposed architecture is shown in Fig.~\ref{fig:net}.

We present a diffusion-based low-light image enhancement framework with explicit illumination modeling as shown in Fig.~\ref{fig:net}. The architecture employs four feature extractors to decompose input low-light images $\mathcal{I}$ into illumination $T_{ref}$, color invariance features $\Phi(x)$, illumination-invariant features $R_c$, and shadow priors $S_{3ch}$.

% We present a diffusion-based low-light image enhancement framework with explicit illumination modeling as shown in Fig.~\ref{fig:net}. The architecture employs a four-branch feature extractor to decompose input images into illumination $T_{ref}$, illumination-invariant features $R_c$, shadow priors $S_{3ch}$, and color invariance $\Phi(x)$. During training, these features are cross-modally concatnated with noisy intermediate states ($X_t$) to guide the diffusion model in learning illumination-aware denoising trajectories. In inference, Gaussian noise ($X_T$) is progressively denoised through feature fusion modules that preserve semantic coherence, generating physically accurate enhanced outputs ($\hat{X}_0$). By explicitly modeling illumination-reflection separation, \sce overcomes traditional diffusion models' single-noise dependency, resolving shadow artifacts and color distortion in low-light conditions.

\subsection{Structured Control Embedding Module}

In this section, we present our method for decomposing a low-light image into its illumination and illumination-invariant information components by leveraging anisotropic structure priors and a Laplacian-regularized optimization framework. Our approach comprises two main stages. First, an illumination refinement procedure is applied using texture-aware weighting and frequency-domain regularization. Second, a post-processing step extracts shadow information from the intermediate results.

Let low-light image $\mathbf{I} \in \mathbb{R}^{H \times W \times 3}$ denote an RGB image captured under non-uniform low-light conditions. We assume that the image is formed as:
\begin{equation}
    \mathbf{I}(x,y) = \mathbf{R}(x,y) \circ \mathbf{L}(x,y)
\end{equation}
where $\mathbf{R}(x,y)$ is the intrinsic illumination-invariant information and $\mathbf{L}(x,y)$ is the illumination map; here, $\circ$ denotes element-wise multiplication.

\textbf{Illumination and Illumination-invariant feature.} An initial illumination estimate is computed by taking the maximum response over the three color channels:
\begin{equation}
    T_{\mathrm{ini}}(x,y) = \max_{c \in \{R,G,B\}}\, \mathbf{I}_c(x,y) + \delta, \quad \delta=0.02
\end{equation}
This initialization guarantees stability even in darker regions.
To preserve structural details while enforcing smoothness, we compute anisotropic weights based on local gradient statistics. The horizontal and vertical finite-difference operators are defined by $f_1$ and $f_2$, which approximate $\partial_x$ and $\partial_y$, respectively. We then obtain the gradients:

{\footnotesize
\begin{equation}
    \nabla_x T_{\mathrm{ini}}(x,y) = (f_1 \bullet T_{\mathrm{ini}})(x,y) = T_{\mathrm{ini}}(x,y+1) - T_{\mathrm{ini}}(x,y)
\end{equation}}
{\footnotesize
\begin{equation}
    \nabla_y T_{\mathrm{ini}}(x,y) = (f_2 \bullet T_{\mathrm{ini}})(x,y) =  T_{\mathrm{ini}}(x+1,y) - T_{\mathrm{ini}}(x,y)
\end{equation}}

where $\bullet$ denotes discrete convolution with reflective boundary handling. A global texture weight is computed as:

{\tiny
\begin{equation}
    w_{\mathrm{to}}(x,y) = \left( \max\left\{\frac{1}{3}\sum_{c=1}^{3}\sqrt{\left(\nabla_x T_{\mathrm{ini},c}(x,y)\right)^2+\left(\nabla_y T_{\mathrm{ini},c}(x,y)\right)^2},\, \varepsilon_s \right\}\right)^{-1}
\end{equation}}
with $\varepsilon_s=0.02$.

To refine the local structure, we smooth $T_{\mathrm{ini}}$ by applying a separable Gaussian filter $\mathcal{G}_{\sigma}$ with standard deviation $\sigma$, resulting in the smoothed image $\hat{T}(x,y) = \mathcal{G}_{\sigma} * T_{\mathrm{ini}}(x,y)$.

% \begin{equation}
%     \hat{T}(x,y) = \mathcal{G}_{\sigma} * T_{\mathrm{ini}}(x,y)
% \end{equation}
The smoothed gradients $\widehat{\nabla}_x \hat{T}$ and $\widehat{\nabla}_y \hat{T}$ yield local weights:
\begin{equation}
    w_{x,\mathrm{local}}(x,y) = \left( \max\left\{\frac{1}{3}\sum_{c=1}^{3}\left|\widehat{\nabla}_{x,c}\hat{T}(x,y)\right|,\, \varepsilon \right\}\right)^{-1}
\end{equation}
\begin{equation}
    w_{y,\mathrm{local}}(x,y) = \left( \max\left\{\frac{1}{3}\sum_{c=1}^{3}\left|\widehat{\nabla}_{y,c}\hat{T}(x,y)\right|,\, \varepsilon \right\}\right)^{-1}
\end{equation}
where $\varepsilon=0.001$. The final anisotropic weights become
\begin{align}
    w_x(x,y) = w_{\mathrm{to}}(x,y) \cdot w_{x,\mathrm{local}}(x,y)
    \notag
    \\ w_y(x,y) = w_{\mathrm{to}}(x,y) \cdot w_{y,\mathrm{local}}(x,y)
\end{align}

To refine the illumination, we solve the following energy minimization problem:

{\footnotesize
\begin{equation}
    E(T) = \|T - T_{\mathrm{ini}}\|_2^2 + \lambda \left(\|\nabla_x T \cdot w_x\|_2^2 + \|\nabla_y T \cdot w_y\|_2^2\right)
\end{equation}}

with regularization parameter $\lambda$. The corresponding Euler-Lagrange equation is discretized to yield the linear system $\left(\mathbf{I} + \lambda\, \mathbf{L}_w\right) \mathbf{t} = \mathbf{b}$,
% \begin{equation}
%     \left(\mathbf{I} + \lambda\, \mathbf{L}_w\right) \mathbf{t} = \mathbf{b}
% \end{equation}
where $\mathbf{t}$ represents the vectorized refined illumination $T_{\mathrm{ref}}$, and $\mathbf{b}$ is computed from $T_{\mathrm{ini}}$, and $\mathbf{L}_w$ is an anisotropic Laplacian, where each pixel $(x,y)$ is connected only to its four immediate neighbors through direction-specific weights $w_x$ and $w_y$ The diagonal entries of $\mathbf{L}_w$ satisfy the following condition:

{\scriptsize
\begin{equation}
    D(x,y) = 1 - \Big(w_x^{\text{right}}(x,y) + w_x^{\text{left}}(x,y) + w_y^{\text{up}}(x,y) + w_y^{\text{down}}(x,y)\Big)
\end{equation}}
% After solving for $T_{\mathrm{ref}}$, a gamma correction is applied:
Since the system described above is entirely linear, it does not account for gamma correction. After obtaining $T_{\mathrm{ref}}$, a pointwise gamma transformation is applied to remap the dynamic range. This nonlinear adjustment is used solely to enhance contrast in the final illumination map.
\begin{equation}
    T_{\mathrm{ref}}(x,y) \leftarrow \left(T_{\mathrm{ref}}(x,y)\right)^{1/\gamma}
\end{equation}
$T_{\mathrm{ref}}$ is our illumination information for the model. 

The final enhanced illumination-invariant information $\mathbf{R}$ is then recovered by:
\begin{equation}
    \mathbf{R}_c(x,y) = \frac{\mathbf{I}_c(x,y)}{T_{\mathrm{ref}}(x,y)}, \quad c \in \{R,G,B\}
\end{equation}

\textbf{Shadow priors.} The shadow information is extracted using the shadow extraction (SE) module. The SE module, illustrated in the gray box in Fig.~\ref{fig:net}, operates as follows:

% After obtaining the illumination‐invariant image $\mathbf{R}$, we extract shadow information via a dedicated shadow extraction (SE) module as shown in the gray box in Fig.~\ref{fig:net}. We then amplify $\mathbf{R}$ by a factor of two, $\mathbf{R}' = 2\,\mathbf{R}$, and invoke the SE function: $S_{3ch} = \mathrm{SE}\bigl(\mathbf{R}')$, $S_{3ch}$ is interpreted as the shadow information. 

% \begin{equation}
%     S_2 = \mathbf{I} - S_1
% \end{equation}
We adopt a frequency-domain strategy based on a discrete Laplacian operator $f_3$. Let $\widehat{f_3}(u,v)$ be its 2D Fourier transform. Additionally, denote by $\widehat{f_1}(u,v)$ and $\widehat{f_2}(u,v)$ the Fourier transforms of $f_1$ and $f_2$ ($f_1$, $f_2$ as previously described), respectively. The frequency-domain update $\kappa$ is then obtained by:

{\scriptsize
\begin{equation}
    \mathcal{F}\{2\mathbf{R}\}(u,v) = \frac{\lambda |\widehat{f_3}(u,v)|^2\, \mathcal{F}\{\mathbf{2R}\}(u,v) + \beta\, \mathcal{F}\{\mathcal{N}_2\}(u,v)}{\lambda |\widehat{f_3}(u,v)|^2 + \beta \Big(|\widehat{f_1}(u,v)|^2 + |\widehat{f_2}(u,v)|^2\Big) + \varepsilon}
\end{equation}}
where $\beta$ is an iteration-dependent parameter (typically defined as $\beta=2^{(i-1)}/\tau$ for threshold $\tau$), $\mathcal{N}_2$ aggregates the soft-thresholded gradient residuals of $2\mathbf{R}$ ---where the factor $2$ serves as an amplification coefficient, and $\varepsilon$ ensures numerical stability. We apply the inverse Fourier transform to the frequency-domain update $\kappa$ to obtain the spatial-domain projection $S_1$, where $S_1(x,y,c) \in [lb(x,y,c),\, ub(x,y,c)]$, with $lb$ and $ub$ representing the lower and upper bounds, respectively.

Thus, we decompose $2\mathbf{R}$ into a smooth structural component $S_1$ and a residual component $S_2$, where $S_2 = 2\mathbf{R} - S_1$. Finally, the output of the SE module is obtained after replicating $S_2$ across three channels: $S_{\mathrm{3ch}} = \mathrm{repmat}(S_2,\,1,\,1,\,3)\in\mathbb{R}^{H\times W\times3}$, ensuring compatibility with standard RGB-based processing.

\textbf{Color invariance.} In order to make the generated luminance-enhanced image stable and invariant in color space; we construct a color representation invariant to global scaling of intensity, which is also a key requirement in tasks involving illumination decomposition, low-light enhancement, or intrinsic image recovery. We define a channel-wise affine-invariant mapping:
\begin{equation}
    \Phi(x) = \left[ \frac{x_r}{\|x_r\|_\infty},\, \frac{x_g}{\|x_g\|_\infty},\, \frac{x_b}{\|x_b\|_\infty} \right]
\end{equation}
where \( \|x_c\|_\infty = \max_{(i,j)} x_c(i,j) \) denotes the channel-wise \(\ell_\infty\)-norm, i.e., the maximum pixel intensity within each color channel.

The mapping \( \Phi \) satisfies the following affine-invariance property:
\begin{equation}
    \Phi(\alpha \cdot x) = \Phi(x),\quad \forall\, \alpha \in \mathbb{R}^{+}
\end{equation}
where $\alpha$ denotes a global positive scalar factor on the image $x$, applied uniformly for each pixel value of the image. Therefore exhibits scale invariance under global illumination changes. In geometric terms, $\Phi(x)$ projects the color vector at each pixel onto the canonical chromaticity subspace, removing dependency on per-channel intensity scaling.

In summary, our method leverages both spatial and frequency domain techniques to effectively separate the illumination and illumination-invariant layers in low-light images while also extracting complementary shadow information. We also extracted color invariant information to maintain color stability. The extracted information was utilized as a conditioning mechanism to control the noise process within the Diffusion model.

\subsection{Fundamental Diffusion}

Our approach leverages a diffusion framework built upon a U-Net backbone to tackle the low-light image enhancement problem. In our model, the forward diffusion process progressively adds Gaussian noise to a clean image $x_0$ over $T$ timesteps. Specifically, at each timestep $t$, the noisy image $x_t$ is defined as
\begin{equation}
    x_t = \sqrt{\bar{\alpha}_t} \, x_0 + \sqrt{1 - \bar{\alpha}_t} \, \epsilon_t,\quad \epsilon_t \sim \mathcal{N}(\mathbf{0},\mathbf{I})
\end{equation}
where \(\bar{\alpha}_t = \prod_{s=1}^{t} \alpha_s\) and each \(\alpha_s \in (0,1)\) controls the noise schedule.

The reverse process aims to reconstruct $x_0$ from $x_t$ by learning to predict the added noise. We parameterize this noise prediction with a U-Net architecture, which effectively preserves high-frequency details through its symmetric encoder-decoder structure and skip connections. Our network is conditioned on auxiliary information, including illumination, illumination-invariant features, shadow cues, and color invariance, which are concatenated to the input.

By incorporating the conditional signals into the diffusion process, our network is guided not only to denoise but also to enhance the relevant structures in low-light scenarios. In summary, our diffusion model with a U-Net backbone offers an effective mechanism for high-fidelity low-light image restoration, ensuring both pixel-level accuracy and perceptual coherence.

% The training objective is based on the simplified loss function~\cite{ho2020denoising, kingma2021variational}, denoted as $L_{\text{simple}}$, which directly measures the discrepancy between the true noise $\epsilon$ and its prediction $\epsilon_{\theta}(x_t, t, c)$:
% \begin{equation}
%     \mathcal{L_{\text{simple}}} = \mathbb{E}_{t,x_0,\epsilon, c} \left[ \left\| \epsilon - \epsilon_{\theta}(x_t, t, c) \right\|_2^2 \right]
% \end{equation}
% where $c$ represents the conditional input comprising the aforementioned auxiliary cues. This formulation stabilizes training and helps the model learn to invert the diffusion process more precisely.

% By incorporating the conditional signals into the diffusion process, our network is guided not only to denoise but also to enhance the relevant structures in low-light scenarios. In summary, our diffusion model with a U-Net backbone and the $L_{\text{simple}}$loss offers an effective mechanism for high-fidelity low-light image restoration, ensuring both pixel-level accuracy and perceptual coherence.

\begin{table*}[htbp]
\centering
\caption{Quantitative evaluation on three datasets: LOLv1~\cite{wei2018deep}, LOLv2-real~\cite{yang2021sparse}, and LSRW~\cite{hai2023r2rnet}. All values are formatted with three decimal places. Higher PSNR and SSIM indicate better performance, while lower LPIPS and FID are preferable. Note that  methods are trained
only on the LOLv1 training set and are directly evaluated on the remaining
datasets}
\label{tab:results}
\resizebox{\textwidth}{!}{%
\begin{tabular}{l l  c c c c  c c c c  c c c c}
\toprule
\multirow{2}{*}{\textbf{Methods}} & \multirow{2}{*}{\textbf{Reference}} & \multicolumn{4}{c}{\textbf{LOL v1}} & \multicolumn{4}{c}{\textbf{LOLv2-real}} & \multicolumn{4}{c}{\textbf{LSRW}} \\
\cmidrule(lr){3-6} \cmidrule(lr){7-10} \cmidrule(lr){11-14}
 &  & PSNR$\uparrow$ & SSIM$\uparrow$ & LPIPS$\downarrow$ & FID$\downarrow$ & PSNR$\uparrow$ & SSIM$\uparrow$ & LPIPS$\downarrow$ & FID$\downarrow$ & PSNR$\uparrow$ & SSIM$\uparrow$ & LPIPS$\downarrow$ & FID$\downarrow$ \\
\midrule
NPE~\cite{wang2013naturalness}             & TIP’ 13    & 16.970 & 0.484 & 0.400 & 104.057 & 17.333 & 0.464 & 0.396 & 100.025 & 16.188 & 0.384 & 0.440 & 90.132 \\
SRIE~\cite{fu2016weighted}            & CVPR’ 16   & 11.855 & 0.495 & 0.353 & 88.728  & 14.451 & 0.524 & 0.332 & 78.834  & 13.357 & 0.415 & 0.399 & 69.082 \\
LIME~\cite{guo2016lime}            & TIP’ 16    & 17.546 & 0.531 & 0.387 & 117.892 & 17.483 & 0.505 & 0.428 & 118.171 & 17.342 & 0.520 & 0.471 & 75.595 \\
RetinexNet~\cite{wei2018deep}      & BMVC’ 18   & 16.774 & 0.462 & 0.417 & 126.266 & 17.715 & 0.652 & 0.436 & 133.905 & 15.609 & 0.414 & 0.454 & 108.350 \\
DSLR~\cite{lim2020dslr}            & TMM’ 20    & 14.816 & 0.572 & 0.375 & 104.428 & 17.000 & 0.596 & 0.408 & 114.306 & 15.259 & 0.441 & 0.464 & 84.930 \\
DRBN~\cite{yang2020fidelity}            & CVPR’ 20   & 16.677 & 0.730 & 0.345 & 98.732  & 18.466 & 0.768 & 0.352 & 89.085  & 16.734 & 0.507 & 0.457 & 80.727 \\
Zero-DCE~\cite{guo2020zero}        & CVPR’ 20   & 14.861 & 0.562 & 0.372 & 87.238  & 18.059 & 0.580 & 0.352 & 80.449  & 15.867 & 0.443 & 0.411 & 63.320 \\
MIRNet~\cite{zamir2020learning}          & ECCV’ 20   & 24.138 & 0.830 & 0.250 & 69.179  & 20.020 & 0.820 & 0.233 & 49.108  & 16.470 & 0.477 & 0.430 & 93.811 \\
EnlightenGAN~\cite{jiang2021enlightengan}    & TIP’ 21    & 17.606 & 0.653 & 0.372 & 94.704  & 18.676 & 0.678 & 0.364 & 84.044  & 17.106 & 0.463 & 0.406 & 69.033 \\
ReLLIE~\cite{zhang2021rellie}          & ACM MM’ 21 & 11.437 & 0.482 & 0.375 & 95.510  & 14.400 & 0.536 & 0.334 & 79.838  & 13.685 & 0.422 & 0.404 & 65.221 \\
RUAS~\cite{liu2021retinex}            & CVPR’ 21   & 16.405 & 0.503 & 0.364 & 101.971 & 15.351 & 0.495 & 0.395 & 94.162  & 14.271 & 0.461 & 0.501 & 78.392 \\
DDIM~\cite{song2020denoising}            & ICLR’ 21   & 16.521 & 0.776 & 0.376 & 84.071  & 15.280 & 0.788 & 0.387 & 76.387  & 14.858 & 0.486 & 0.495 & 71.812 \\
CDEF~\cite{lei2022low}            & TMM’ 22    & 16.335 & 0.585 & 0.407 & 90.620  & 19.757 & 0.630 & 0.349 & 74.055  & 16.758 & 0.465 & 0.399 & 62.780 \\
SCI~\cite{ma2022toward}             & CVPR’ 22   & 14.784 & 0.525 & 0.366 & 78.598  & 17.304 & 0.540 & 0.345 & 67.624  & 15.242 & 0.419 & 0.404 & \underline{56.261} \\
URetinex-Net~\cite{wu2022uretinex}    & CVPR’ 22   & 19.842 & 0.824 & 0.237 & 52.383  & 21.093 & 0.858 & 0.208 & 49.836  & 18.271 & 0.518 & 0.419 & 66.871 \\
SNRNet~\cite{xu2022snr}          & CVPR’ 22   & 24.610 & 0.842 & 0.233 & 55.121  & 21.480 & 0.849 & 0.237 & 54.532  & 16.499 & 0.505 & 0.419 & 65.807 \\
Uformer~\cite{wang2022uformer}         & CVPR’ 22   & 19.001 & 0.741 & 0.354 & 109.351 & 18.442 & 0.759 & 0.347 & 98.138  & 16.591 & 0.494 & 0.435 & 82.299 \\
Restormer~\cite{zamir2022restormer}       & CVPR’ 22   & 20.614 & 0.797 & 0.288 & 72.998  & 24.910 & 0.851 & 0.264 & 58.649  & 16.303 & 0.453 & 0.427 & 69.219 \\
Palette~\cite{saharia2022palette}         & SIGGRAPH’22& 11.771 & 0.561 & 0.498 & 108.291 & 14.703 & 0.692 & 0.333 & 83.942  & 13.570 & 0.476 & 0.479 & 73.841 \\
UHDFour2$\times$~\cite{li2023embedding} & ICLR’ 23   & 23.093 & 0.821 & 0.259 & 56.912  & 21.785 & 0.854 & 0.292 & 60.837  & 17.300 & 0.529 & 0.443 & 62.032 \\
WeatherDiff~\cite{ozdenizci2023restoring}      & TPAMI’ 23  & 17.913 & 0.811 & 0.272 & 73.903  & 20.009 & 0.829 & 0.253 & 59.670  & 16.507 & 0.487 & 0.431 & 96.050 \\
GDP~\cite{fei2023generative}             & CVPR’ 23   & 15.896 & 0.542 & 0.421 & 117.456 & 14.290 & 0.493 & 0.435 & 102.416 & 12.887 & 0.362 & 0.412 & 76.908 \\
DiffLL~\cite{jiang2023low}           & ACM TOG'23 & \underline{26.336} & \underline{0.845} & \underline{0.217} & \underline{48.114}  & \underline{28.857} & \underline{0.876} & \underline{0.207} & \underline{45.359}  & \underline{19.281} & \underline{0.552} & \underline{0.350} & \textbf{45.294} \\
QuadPrior~\cite{wang2024zero} & CVPR' 24& 22.849 &  0.800 &  0.201 & $-$ & 20.592 &  0.811 &  0.202 & $-$ & $-$ & $-$ & $-$ & $-$ \\
Lightendiffusion~\cite{jiang2024lightendiffusion} & ECCV' 24& 20.188 &   0.814 &  0.316 &  85.930 &  22.697 &  0.853 &  0.306 & 75.582 &  18.397 &  0.534 & 0.428 & 67.801 \\
\textbf{Ours}            & $-$         & \textbf{26.947} & \textbf{0.921} & \textbf{0.071} & \textbf{46.670}  & \textbf{31.223} & \textbf{0.926} & \textbf{0.044} & \textbf{36.875}  & \textbf{20.692} & \textbf{0.560} & \textbf{0.198} & 65.710 \\
\bottomrule
\end{tabular}%
}
\vspace{-2mm}
\end{table*}

\subsection{Loss}

The loss function setting in this paper mainly refers to the loss function design of CLEDiffusion~\cite{yin2023cle}.

\textbf{Diffusion Loss.} The training objective is based on the simplified loss function~\cite{ho2020denoising, kingma2021variational}, denoted as $L_{\text{simple}}$, which directly measures the discrepancy between the true noise $\epsilon$ and its prediction $\epsilon_{\theta}(x_t, t, c)$:
\begin{equation}
    \mathcal{L_{\text{simple}}} = \mathbb{E}_{t,x_0,\epsilon, c} \left[ \left\| \epsilon - \epsilon_{\theta}(x_t, t, c) \right\|_2^2 \right]
\end{equation}
where $c$ represents the conditional input comprising the aforementioned auxiliary cues. This formulation stabilizes training and helps the model learn to invert the diffusion process more precisely.

\textbf{Illumination Alignment Loss.} To enforce global brightness consistency, we define
\begin{equation}
    \mathcal{L}_{\mathrm{illum}} = \left\lVert G(\hat{x}_0) - G(x_0) \right\rVert_{1}
\end{equation}
where \(G(\cdot)\) converts an RGB image to grayscale. This loss ensures that the enhanced output maintains similar overall luminance to the ground truth.

\textbf{Chromatic Fidelity Loss.} To reduce color distortion, we minimize the angular difference between normalized RGB vectors:
\begin{equation}
    \mathcal{L}_{\mathrm{chrom}} = \sum_{i=1}^{H \times W}\left(1 - \frac{\hat{x}_0^{(i)} \cdot x_0^{(i)}}{\|\hat{x}_0^{(i)}\|_2 \, \|x_0^{(i)}\|_2}\right)
\end{equation}
where the index \(i\) runs over all pixels, promoting accurate chromatic alignment.

\textbf{Structural Similarity Loss.} To preserve local texture and structure, we adopt an SSIM-based loss:

\begin{equation}
    \mathcal{L}_{\mathrm{SSIM}} = 1 - \frac{(2\mu_{x_0}\mu_{\hat{x}_0} + c_1)(2\sigma_{x_0\hat{x}_0} + c_2)}{(\mu_{x_0}^2 + \mu_{\hat{x}_0}^2 + c_1)(\sigma_{x_0}^2 + \sigma_{\hat{x}_0}^2 + c_2)}
\end{equation}

with $\mu$ and $\sigma$ denoting local means and standard deviations (computed over a sliding window), $\sigma_{x_0\hat{x}_0}$ the covariance, and constants $c_1, c_2$ ensuring numerical stability.

\textbf{Deep Feature Consistency Loss.} To further align high-level semantic details, we employ a VGG-based perceptual loss:
\begin{equation}
    \mathcal{L}_{\mathrm{feat}} = \sum_{l \in L} \frac{1}{N_l} \left\lVert \phi_V^{l}(\hat{x}_0) - \phi_V^{l}(x_0) \right\rVert_{2}^2
\end{equation}
where \(\phi_V^{l}(\cdot)\) extracts features from the \(l\)-th layer of VGGNet and \(N_l = H_l \times W_l \times C_l\) normalizes for feature dimension.

\textbf{Total Loss :} The total loss is a weighting of the above losses:
\begin{align}
    \mathcal{L_{\text{Total}}} = &\mathcal{L_{\text{simple}}} + \omega_{\text{illum}}\mathcal{L}_{\text{illum}} + \omega_{\text{chrom}}\mathcal{L}_{\text{chrom}}
    \notag
    \\ &+ \omega_{\text{SSIM}}\mathcal{L}_{\text{SSIM}} + \omega_{\text{feat}}\mathcal{L_{\text{feat}}}
\end{align}
where $\omega_{\text{illum}}$, $\omega_{\text{chrom}}$, $\omega_{\text{ssim}}$, $\omega_{\text{feat}}$ are the weights of the corresponding losses and we use the same values as in \cite{yin2023cle}.

%%%%%%%%%%%%%%%%%%%%%%%%%%%%%%%%%%%%%%%%%%%%%%%%%%%%%%%%%%%%%%%%%%%%%%%%%%%%%%%%%%%%%%%%%%%%%%%%%%%%%%%%%%%%%%%%%%%%%%%%%%%%%%%%%%%%%%%%%%%%%%%
\section{EXPERIMENT}

\begin{figure*}[h]
  % \vspace{-1cm}
  \centering
  \includegraphics[width=0.87\linewidth]{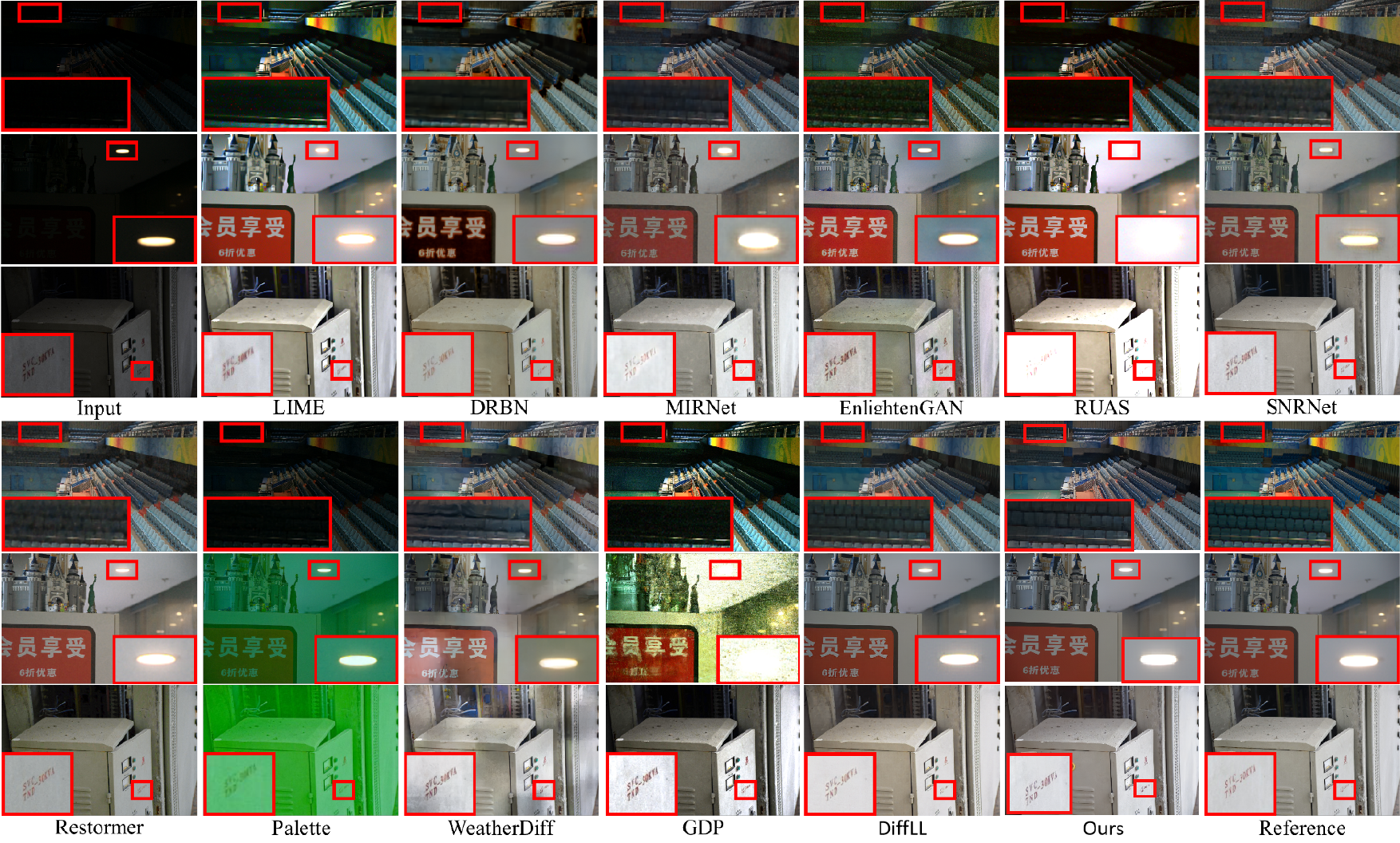}
  \caption{Visual comparisons of our approach with competing methods. The input image is from datasets LOLv1, LOLv2-real, LSRW for the first, second and third rows, respectively.}
  \vspace{-4mm}
  \label{fig:comper}
\end{figure*}

\subsection{Experimental Settings}

\textbf{Implementation Details.} Our models are trained on NVIDIA A40 GPUs. The AdamW~\cite{loshchilov2017decoupled} optimizer was used and the learning rate was set to $5 \times 10^{-5}$ with a learning rate decay coefficient of $1 \times 10^{-4}$.  The batch size and patch size are set to 8 and $256 \times 256$ respectively. The backbone network of Diffusion is the commonly used U-Net network structure~\cite{ronneberger2015u}. The time step $T$ is set to 1000 for the training phase and the implicit sampling step is set to 100 for both the training and inference phases.

\textbf{Datasets.} Our model is trained on the LOLv1~\cite{wei2018deep} real image dataset. The LOLv1 real image dataset contains 500 pairs of real images, of which 485 pairs are used for training and 15 pairs are used for evaluation. We also evaluated our model directly on the real image portion of the LOLv2~\cite{yang2021sparse} dataset as well as the LSRW~\cite{hai2023r2rnet} dataset without training and fine-tuning. Our model is tested directly on 100 pairs of evaluation images for LOLv2-real and 50 pairs of evaluation images for LSRW. In addition to verify the generalization ability of our model we add tests on the DICM dataset~\cite{lee2013contrast}, the MEF dataset~\cite{ma2015perceptual}, and the LIME dataset~\cite{guo2016lime}.

\textbf{Metrics.} In our experiments we use two evaluation metrics commonly used in low-light enhancement tasks, PSNR and SSIM~\cite{wang2004image}, for evaluation. We also use two perceptual evaluation metrics LPIPS~\cite{zhang2018unreasonable} as well as FID~\cite{heusel2017gans} to evaluate the images. In this paper LPIPS uses AlexNet~\cite{krizhevsky2012imagenet} as the backbone. In verifying the generalization ability of the model, we use NIQE~\cite{mittal2012making}, BRISQUE~\cite{mittal2012no} and PI~\cite{blau20182018} as evaluation metrics for those datasets that do not contain paired images.

\subsection{Comparison Methods}

\textbf{Quantitative Comparison.} To validate our diffusion‐based enhancement framework with structured illumination priors, we performed experiments on LOLv1, LOLv2‑real and LSRW (Table~\ref{tab:results}), training only on LOLv1 and evaluating off‐the‐shelf elsewhere. Our method outperforms prior work on all metrics and datasets. On LOLv1, it achieves PSNR 26.947 and SSIM 0.921—surpassing DiffLL (26.336/0.845) and SNRNet (24.610/0.842)—and sets a new LPIPS record of 0.071 (vs.\ 0.201). On LOLv2‑real, it generalizes strongly with PSNR 31.223 and SSIM 0.926 (+2.366 dB, +0.050 vs.\ DiffLL), while securing the lowest FID 36.875 and LPIPS 0.044. On LSRW, despite diverse scenes and lighting, our framework maintains the best PSNR 20.692, LPIPS 0.198, and SSIM 0.560; being still competitive on FID. Thus, our perceptual consistency, driven by explicit illumination, shadow, and color‐invariant conditioning, is superior to previous SOTA methods.

% In Table~\ref{tab:uhdll_results}, we evaluated the low-light enhancement capability of our model on high-resolution images. Our method achieves the highest PSNR (27.166), SSIM (0.950), and lowest LPIPS (0.074) on the UHD-LL test set, while its FID (34.868) is competitive with state-of-the-art methods. Notably, despite operating on inputs at only 720×480 resolution, our approach—leveraging illumination decomposition and noise-conditioned diffusion—demonstrates superior structure restoration and perceptual fidelity, significantly outperforming models trained solely on LOLv1. These results highlight the strong generalization capabilities of our method across resolutions and datasets.

As shown in Table~\ref{tab:quality_results}, we validate the generalization ability of our proposed model. Our method achieves the best perceptual quality on DICM across all metrics (NIQE, BRI, PI), on LIME for both BRI and PI, and on MEF in terms of BRI. Moreover, our approach ranks second in terms of NIQE on LIME. Compared with DiffLL, which consistently performs well, our approach further reduces distortion and artifacts, delivering sharper, more natural-looking results. These findings demonstrate the robustness and generalization ability of our method across diverse low-light scenarios.

The results validate the robustness and superiority of our method. By integrating illumination decomposition, shadow priors, and color constraints into the conditional diffusion process, we effectively guide the denoising trajectory and facilitate the generation of visually plausible and structurally faithful enhanced images.
\begin{table}[t!]
\centering
\caption{
Quantitative comparison of different methods on the DICM~\cite{lee2013contrast}, MEF~\cite{ma2015perceptual}, and LIME~\cite{guo2016lime} datasets.
The best results are highlighted in \textbf{bold} and the second-best in \underline{underlined}, and BRI represents BRISQUE~\cite{mittal2012no}.}
\label{tab:quality_results}
\resizebox{\linewidth}{!}{
\begin{tabular}{lccccccccc}
\toprule
\multirow{2}{*}{\textbf{Methods}} & \multicolumn{3}{c}{\textbf{DICM}} & \multicolumn{3}{c}{\textbf{MEF}} & \multicolumn{3}{c}{\textbf{LIME}} \\
\cmidrule(lr){2-4} \cmidrule(lr){5-7} \cmidrule(lr){8-10}
 & {NIQE$\downarrow$} & {BRI$\downarrow$} & {PI$\downarrow$} & {NIQE$\downarrow$} & {BRI$\downarrow$} & {PI$\downarrow$} & {NIQE$\downarrow$} & {BRI$\downarrow$} & {PI$\downarrow$} \\
\midrule
$\text{LIME}_{\text{\cite{guo2016lime}}}$             & 4.476 & 27.375 & 4.216 & 4.744 & 39.095 & 5.160 & 5.045 & 32.842 & 4.859 \\
$\text{DRBN}_{\text{\cite{yang2020fidelity}}}$          & 4.369 & 30.708 & 3.800 & 4.869 & 44.669 & 4.711 & 4.562 & 34.564 & 3.973 \\
$\text{Zero-DCE}_{\text{\cite{guo2020zero}}}$         & 3.951 & 23.350 & 3.149 & \underline{3.500} & 29.359 & \textbf{2.989} & 4.379 & 26.054 & 3.239 \\
$\text{MIRNet}_{\text{\cite{zamir2020learning}}}$           & 4.021 & 22.104 & 3.691 & 4.202 & 34.499 & 3.756 & 4.378 & 28.623 & 3.398 \\
$\text{EnlightenGAN}_{\text{\cite{jiang2021enlightengan}}}$     & 3.832 & 19.129 & 3.256 & 3.556 & 26.799 & 3.270 & 4.249 & 22.664 & 3.381 \\
$\text{RUAS}_{\text{\cite{liu2021retinex}}}$             & 7.306 & 46.882 & 5.700 & 5.435 & 42.120 & 4.921 & 5.322 & 34.880 & 4.581 \\
$\text{DDIM}_{\text{\cite{song2020denoising}}}$             & 3.899 & 19.787 & 3.213 & 3.621 & 28.614 & 3.376 & 4.399 & 24.474 & 3.459 \\
$\text{SCI}_{\text{\cite{ma2022toward}}}$            & 4.519 & 27.922 & 3.700 & 3.608 & 26.716 & 3.286 & 4.463 & 25.170 & 3.376 \\
$\text{URetinex-Net}_{\text{\cite{wu2022uretinex}}}$    & 4.774 & 24.544 & 3.565 & 4.231 & 34.720 & 3.665 & 4.694 & 29.022 & 3.713 \\
$\text{SNRNet}_{\text{\cite{xu2022snr}}}$          & 3.804 & 19.459 & 3.285 & 4.063 & 28.331 & 3.753 & 4.597 & 29.023 & 3.677 \\
$\text{Uformer}_{\text{\cite{wang2022uformer}}}$          & 3.847 & 19.657 & 3.180 & 3.935 & 25.240 & 3.582 & 4.300 & 21.874 & 3.565 \\
$\text{Restormer}_{\text{\cite{zamir2022restormer}}}$       & 3.964 & 19.474 & 3.152 & 3.815 & 25.322 & 3.436 & 4.365 & 22.931 & 3.292 \\
$\text{Palette}_{\text{\cite{saharia2022palette}}}$         & 4.118 & 18.732 & 3.425 & 4.459 & 25.602 & 4.205 & 4.485 & 20.551 & 3.579 \\
$\text{UHDFour2$\times$}_{\text{\cite{li2023embedding}}}$ & 4.575 & 26.926 & 3.684 & 4.231 & 29.538 & 4.124 & 4.430 & 20.263 & 3.813 \\
$\text{WeatherDiff}_{\text{\cite{ozdenizci2023restoring}}}$      & \underline{3.773} & 20.387 & 3.130 & 3.753 & 30.480 & 3.312 & 4.312 & 28.090 & 3.424 \\
$\text{GDP}_{\text{\cite{fei2023generative}}}$              & 4.358 & 19.294 & 3.552 & 4.609 & 34.859 & 4.115 & 4.891 & 27.460 & 3.694 \\
$\text{DiffLL}_{\text{\cite{jiang2023low}}}$          & 3.806 & \underline{18.584} & \underline{3.011} & \textbf{3.427} & \underline{24.165} & \underline{3.011} & \textbf{3.777} & \underline{19.843} & \underline{3.074} \\
Ours             &  \textbf{3.355} &  \textbf{16.676} & \textbf{2.606} & 3.827 & \textbf{21.092} & 3.171 & \underline{4.091} & \textbf{16.820} & \textbf{2.838} \\
\bottomrule
\end{tabular}
}
\end{table}

\begin{table}[t!]
\centering
\caption{Ablation study for our proposed SCEM module on LOLv1.}
\vspace{-2mm}
\label{tab:ablation_study}
\resizebox{0.5\linewidth}{!}{
\begin{tabular}{cccc}
\toprule
\textbf{SCEM} & \textbf{PSNR} $\uparrow$ & \textbf{SSIM} $\uparrow$ & \textbf{LPIPS} $\downarrow$ \\
\midrule
$-$             & 22.220 & 0.810 &  0.220\\
\checkmark    & \textbf{26.947} & \textbf{0.921} & \textbf{0.071}\\
\bottomrule
\end{tabular}
}
\end{table}

\textbf{Qualitative Comparison.} Fig.~\ref{fig:comper} shows a qualitative comparison of three different images. Our method recovers both local and global structures under challenging low-light conditions and accurately captures subtle boundaries, restores the fine texture, and mitigates the color shift throughout. In contrast, for the other methods, certain foreground regions appear overenhanced or remain too dim, with incomplete artifact removal and lost edge definition.

\begin{figure}
  % \vspace{-1cm}
  \centering
  \includegraphics[width=0.9\linewidth]{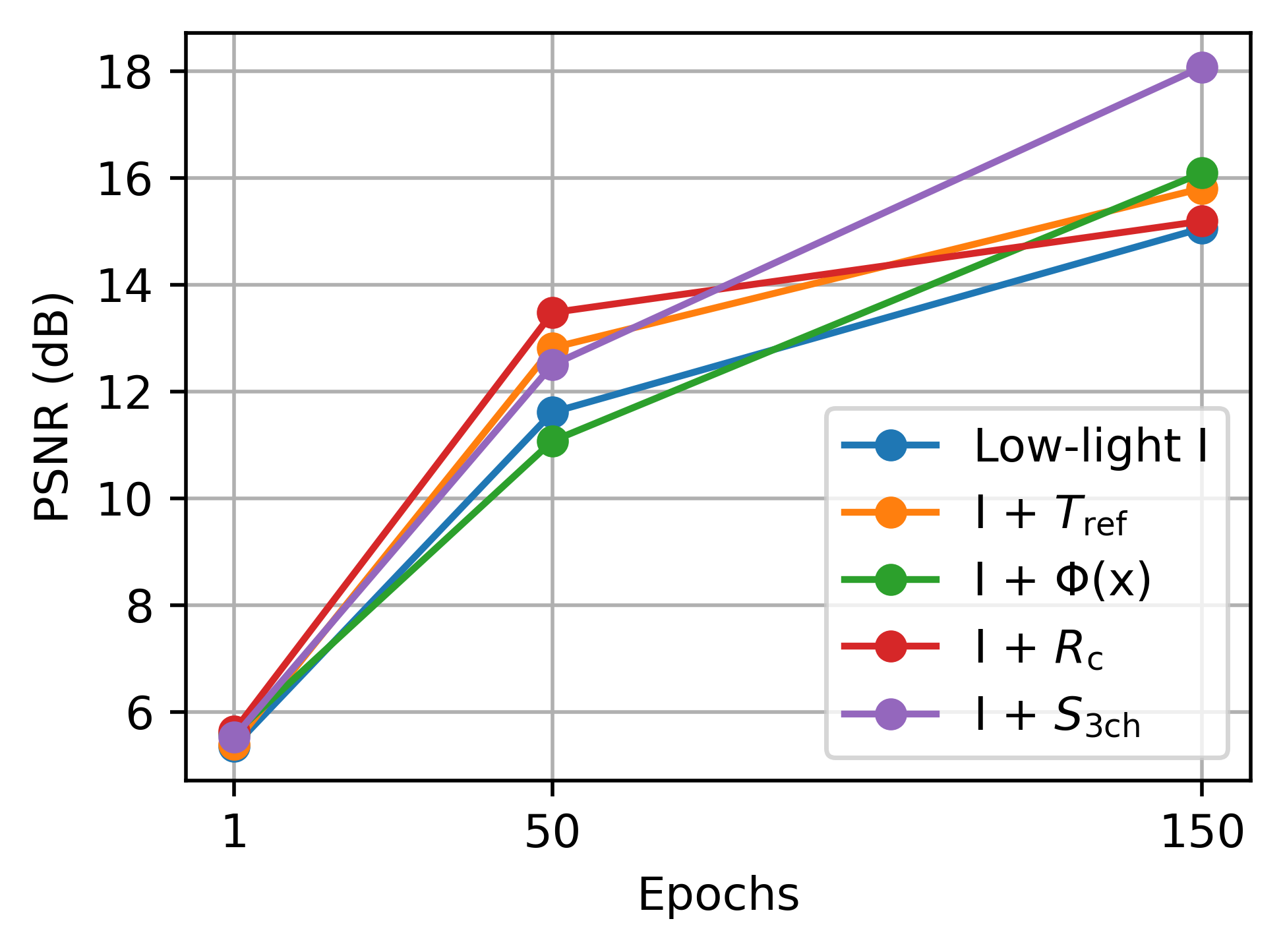}
  \vspace{-2mm}
  \caption{Ablation study comparing PSNR for different input configurations: (1) low-light image $\mathcal{I}$ only, (2) $\mathcal{I}$ with illumination $T_{ref}$, (3) $\mathcal{I}$ with color-invariant features $\Phi(x)$, (4) $\mathcal{I}$ with illumination-invariant features $R_c$, and (5) $\mathcal{I}$ with shadow priors $S_{3ch}$. }
   \vspace{-3mm}
  \label{fig:abs}
\end{figure}

\subsection{Ablation Study}

We utilize LOLv1 for our ablation study, focusing on the four main components of the proposed SCEM: illumination $T_{ref}$, color-invariant features $\Phi(x)$, illumination-invariant features $R_c$, and shadow priors $S_{3ch}$. The default input is the low-light image $\mathcal{I}$, and we evaluate the contribution of each component to the LLIE process by concatenating $\mathcal{I}$ with each component, as illustrated in Fig.~\ref{fig:abs}. All variants show similar performance in the first epoch (PSNR $5.3$–$5.6$ dB, SSIM $0.03$–$0.05$), indicating that auxiliary inputs do not yield significant benefit under extremely limited training.

At epoch 50, the baseline attains 11.61 dB / 0.3414. Adding color‐invariant features slightly degrades performance (11.07 dB / 0.3012), underscoring the stage‐dependent value of different priors. Reference illumination raises PSNR to 12.82 dB (SSIM 0.3328), while illumination‐invariant inputs achieve 13.48 dB / 0.5182. Shadow priors similarly boost SSIM (0.5139) with PSNR 12.50 dB. By epoch 150, the baseline reaches 15.06 dB/0.7035. Color‐invariant features yield the highest PSNR (16.09 dB) at the expense of SSIM (0.6397); illumination cues continue to aid enhancement; and shadow priors deliver the top PSNR (18.08 dB) but moderate SSIM (0.5718).

In summary, shadow priors maximize final PSNR, illumination‐invariant representations most effectively accelerate convergence and structural fidelity, color‐invariant features trade structure for color consistency, and explicit illumination cues provide modest acceleration, guiding the design of hybrid inputs for brightness versus structure objectives.

In Table~\ref{tab:ablation_study}, our ablation study on the LOLv1 dataset demonstrates that the inclusion of the SCEM conditional input dramatically improves the model's performance. Without this conditional mechanism, the model achieves a PSNR of 22.220 dB, SSIM of 0.810, and LPIPS of 0.220. With our proposed SCEM module, PSNR rises to 26.947 dB, SSIM to 0.921, and LPIPS drops to 0.071. These improvements indicate that the conditional mechanism plays a vital role in guiding the enhancement process, leading to significantly lower reconstruction errors, improved structural fidelity, and superior perceptual quality, thereby validating its critical importance in our low-light image enhancement framework.
% \vspace{-0.2cm}
%%%%%%%%%%%%%%%%%%%%%%%%%%%%%%%%%%%%%%%%%%%%%%%%%%%%%%%%%%%%%%%%%%%%%%%%%%%%%%%%%%%%%%%%%%%%%%%%%%%%%%%%%%%%%%%%%%%%%%%%%%%%%%%%%%%%%%%%%%%%%%%
\section{CONCLUSION}
We present a novel diffusion-based low-light image enhancement method that is conditioned on color and luminance priors. More in detail, we propose to use four specific priors: illumination, illumination-invariant, shadow, and color invariance priors. These priors are introduced into the diffusion process thanks to our proposed Structured Control Embedding Module. Experimental results demonstrate that our model achieves state-of-the-art performance and exhibits strong generalization across diverse datasets without fine-tuning. These findings underscore the robustness and effectiveness of our method in restoring visual quality under challenging lighting conditions.

\bibliographystyle{IEEEtran}
\bibliography{icra2026}

@STRING{tog       = "ACM Transactions on Graphics"}

@String{Computing = "Computing" }

@String{Computer = "{IEEE} Computer" }

@String{Springer = "Springer-Verlag" }

@article{li2023embedding,
  title={Embedding fourier for ultra-high-definition low-light image enhancement},
  author={Li, Chongyi and Guo, Chun-Le and Zhou, Man and Liang, Zhexin and Zhou, Shangchen and Feng, Ruicheng and Loy, Chen Change},
  journal={arXiv preprint arXiv:2302.11831},
  year={2023}
}

@inproceedings{xu2023low,
  title={Low-light image enhancement via structure modeling and guidance},
  author={Xu, Xiaogang and Wang, Ruixing and Lu, Jiangbo},
  booktitle={Proceedings of the IEEE/CVF Conference on Computer Vision and Pattern Recognition},
  pages={9893--9903},
  year={2023}
}

@inproceedings{yin2023cle,
  title={Cle diffusion: Controllable light enhancement diffusion model},
  author={Yin, Yuyang and Xu, Dejia and Tan, Chuangchuang and Liu, Ping and Zhao, Yao and Wei, Yunchao},
  booktitle={Proceedings of the 31st ACM International Conference on Multimedia},
  pages={8145--8156},
  year={2023}
}

@article{song2020denoising,
  title={Denoising diffusion implicit models},
  author={Song, Jiaming and Meng, Chenlin and Ermon, Stefano},
  journal={arXiv preprint arXiv:2010.02502},
  year={2020}
}

@article{ho2020denoising,
  title={Denoising diffusion probabilistic models},
  author={Ho, Jonathan and Jain, Ajay and Abbeel, Pieter},
  journal={Advances in neural information processing systems},
  volume={33},
  pages={6840--6851},
  year={2020}
}

@article{yang2018low,
  title={Low-light image enhancement based on Retinex theory and dual-tree complex wavelet transform},
  author={Yang, Mao-xiang and Tang, Gui-jin and Liu, Xiao-hua and Wang, Li-qian and Cui, Zi-guan and Luo, Su-huai},
  journal={Optoelectronics Letters},
  volume={14},
  number={6},
  pages={470--475},
  year={2018},
  publisher={Springer}
}

@article{jeon2024low,
  title={Low-light image enhancement using gamma correction prior in mixed color spaces},
  author={Jeon, Jong Ju and Park, Jun Young and Eom, Il Kyu},
  journal={Pattern Recognition},
  volume={146},
  pages={110001},
  year={2024},
  publisher={Elsevier}
}

@article{land1971lightness,
  title={Lightness and retinex theory},
  author={Land, Edwin H and McCann, John J},
  journal={Journal of the Optical society of America},
  volume={61},
  number={1},
  pages={1--11},
  year={1971},
  publisher={Optical Society of America}
}

@article{jebadass2022low,
  title={Low light enhancement algorithm for color images using intuitionistic fuzzy sets with histogram equalization},
  author={Jebadass, J Reegan and Balasubramaniam, P},
  journal={Multimedia Tools and Applications},
  volume={81},
  number={6},
  pages={8093--8106},
  year={2022},
  publisher={Springer}
}

@inproceedings{wang2023ultra,
  title={Ultra-high-definition low-light image enhancement: A benchmark and transformer-based method},
  author={Wang, Tao and Zhang, Kaihao and Shen, Tianrun and Luo, Wenhan and Stenger, Bjorn and Lu, Tong},
  booktitle={Proceedings of the AAAI conference on artificial intelligence},
  volume={37},
  number={3},
  pages={2654--2662},
  year={2023}
}

@inproceedings{zhou2025low,
  title={Low-light image enhancement via generative perceptual priors},
  author={Zhou, Han and Dong, Wei and Liu, Xiaohong and Zhang, Yulun and Zhai, Guangtao and Chen, Jun},
  booktitle={Proceedings of the AAAI Conference on Artificial Intelligence},
  volume={39},
  number={10},
  pages={10752--10760},
  year={2025}
}

@article{wei2018deep,
  title={Deep retinex decomposition for low-light enhancement},
  author={Wei, Chen and Wang, Wenjing and Yang, Wenhan and Liu, Jiaying},
  journal={arXiv preprint arXiv:1808.04560},
  year={2018}
}

@article{lee2013contrast,
  title={Contrast enhancement based on layered difference representation of 2D histograms},
  author={Lee, Chulwoo and Lee, Chul and Kim, Chang-Su},
  journal={IEEE transactions on image processing},
  volume={22},
  number={12},
  pages={5372--5384},
  year={2013},
  publisher={IEEE}
}

@article{ma2015perceptual,
  title={Perceptual quality assessment for multi-exposure image fusion},
  author={Ma, Kede and Zeng, Kai and Wang, Zhou},
  journal={IEEE Transactions on Image Processing},
  volume={24},
  number={11},
  pages={3345--3356},
  year={2015},
  publisher={IEEE}
}

@article{guo2016lime,
  title={LIME: Low-light image enhancement via illumination map estimation},
  author={Guo, Xiaojie and Li, Yu and Ling, Haibin},
  journal={IEEE Transactions on image processing},
  volume={26},
  number={2},
  pages={982--993},
  year={2016},
  publisher={IEEE}
}

@article{mittal2012no,
  title={No-reference image quality assessment in the spatial domain},
  author={Mittal, Anish and Moorthy, Anush Krishna and Bovik, Alan Conrad},
  journal={IEEE Transactions on image processing},
  volume={21},
  number={12},
  pages={4695--4708},
  year={2012},
  publisher={IEEE}
}

@article{yang2021sparse,
  title={Sparse gradient regularized deep retinex network for robust low-light image enhancement},
  author={Yang, Wenhan and Wang, Wenjing and Huang, Haofeng and Wang, Shiqi and Liu, Jiaying},
  journal={IEEE Transactions on Image Processing},
  volume={30},
  pages={2072--2086},
  year={2021},
  publisher={IEEE}
}

@article{hai2023r2rnet,
  title={R2rnet: Low-light image enhancement via real-low to real-normal network},
  author={Hai, Jiang and Xuan, Zhu and Yang, Ren and Hao, Yutong and Zou, Fengzhu and Lin, Fang and Han, Songchen},
  journal={Journal of Visual Communication and Image Representation},
  volume={90},
  pages={103712},
  year={2023},
  publisher={Elsevier}
}

@article{loshchilov2017decoupled,
  title={Decoupled weight decay regularization},
  author={Loshchilov, Ilya and Hutter, Frank},
  journal={arXiv preprint arXiv:1711.05101},
  year={2017}
}

@inproceedings{ronneberger2015u,
  title={U-net: Convolutional networks for biomedical image segmentation},
  author={Ronneberger, Olaf and Fischer, Philipp and Brox, Thomas},
  booktitle={Medical image computing and computer-assisted intervention--MICCAI 2015: 18th international conference, Munich, Germany, October 5-9, 2015, proceedings, part III 18},
  pages={234--241},
  year={2015},
  organization={Springer}
}

@article{wang2004image,
  title={Image quality assessment: from error visibility to structural similarity},
  author={Wang, Zhou and Bovik, Alan C and Sheikh, Hamid R and Simoncelli, Eero P},
  journal={IEEE transactions on image processing},
  volume={13},
  number={4},
  pages={600--612},
  year={2004},
  publisher={IEEE}
}

@inproceedings{zhang2018unreasonable,
  title={The unreasonable effectiveness of deep features as a perceptual metric},
  author={Zhang, Richard and Isola, Phillip and Efros, Alexei A and Shechtman, Eli and Wang, Oliver},
  booktitle={Proceedings of the IEEE conference on computer vision and pattern recognition},
  pages={586--595},
  year={2018}
}

@article{heusel2017gans,
  title={Gans trained by a two time-scale update rule converge to a local nash equilibrium},
  author={Heusel, Martin and Ramsauer, Hubert and Unterthiner, Thomas and Nessler, Bernhard and Hochreiter, Sepp},
  journal={Advances in neural information processing systems},
  volume={30},
  year={2017}
}

@article{mittal2012making,
  title={Making a “completely blind” image quality analyzer},
  author={Mittal, Anish and Soundararajan, Rajiv and Bovik, Alan C},
  journal={IEEE Signal processing letters},
  volume={20},
  number={3},
  pages={209--212},
  year={2012},
  publisher={IEEE}
}

@inproceedings{blau20182018,
  title={The 2018 PIRM challenge on perceptual image super-resolution},
  author={Blau, Yochai and Mechrez, Roey and Timofte, Radu and Michaeli, Tomer and Zelnik-Manor, Lihi},
  booktitle={Proceedings of the European conference on computer vision (ECCV) workshops},
  pages={0--0},
  year={2018}
}

@article{krizhevsky2012imagenet,
  title={Imagenet classification with deep convolutional neural networks},
  author={Krizhevsky, Alex and Sutskever, Ilya and Hinton, Geoffrey E},
  journal={Advances in neural information processing systems},
  volume={25},
  year={2012}
}

@article{wang2013naturalness,
  title={Naturalness preserved enhancement algorithm for non-uniform illumination images},
  author={Wang, Shuhang and Zheng, Jin and Hu, Hai-Miao and Li, Bo},
  journal={IEEE transactions on image processing},
  volume={22},
  number={9},
  pages={3538--3548},
  year={2013},
  publisher={IEEE}
}

@inproceedings{fu2016weighted,
  title={A weighted variational model for simultaneous reflectance and illumination estimation},
  author={Fu, Xueyang and Zeng, Delu and Huang, Yue and Zhang, Xiao-Ping and Ding, Xinghao},
  booktitle={Proceedings of the IEEE conference on computer vision and pattern recognition},
  pages={2782--2790},
  year={2016}
}

@article{lei2022low,
  title={Low-light image enhancement using the cell vibration model},
  author={Lei, Xiaozhou and Fei, Zixiang and Zhou, Wenju and Zhou, Huiyu and Fei, Minrui},
  journal={IEEE Transactions on Multimedia},
  volume={25},
  pages={4439--4454},
  year={2022},
  publisher={IEEE}
}

@article{lim2020dslr,
  title={DSLR: Deep stacked Laplacian restorer for low-light image enhancement},
  author={Lim, Seokjae and Kim, Wonjun},
  journal={IEEE Transactions on Multimedia},
  volume={23},
  pages={4272--4284},
  year={2020},
  publisher={IEEE}
}

@inproceedings{yang2020fidelity,
  title={From fidelity to perceptual quality: A semi-supervised approach for low-light image enhancement},
  author={Yang, Wenhan and Wang, Shiqi and Fang, Yuming and Wang, Yue and Liu, Jiaying},
  booktitle={Proceedings of the IEEE/CVF conference on computer vision and pattern recognition},
  pages={3063--3072},
  year={2020}
}

@inproceedings{guo2020zero,
  title={Zero-reference deep curve estimation for low-light image enhancement},
  author={Guo, Chunle and Li, Chongyi and Guo, Jichang and Loy, Chen Change and Hou, Junhui and Kwong, Sam and Cong, Runmin},
  booktitle={Proceedings of the IEEE/CVF conference on computer vision and pattern recognition},
  pages={1780--1789},
  year={2020}
}

@inproceedings{zamir2020learning,
  title={Learning enriched features for real image restoration and enhancement},
  author={Zamir, Syed Waqas and Arora, Aditya and Khan, Salman and Hayat, Munawar and Khan, Fahad Shahbaz and Yang, Ming-Hsuan and Shao, Ling},
  booktitle={Computer Vision--ECCV 2020: 16th European Conference, Glasgow, UK, August 23--28, 2020, Proceedings, Part XXV 16},
  pages={492--511},
  year={2020},
  organization={Springer}
}

@article{jiang2021enlightengan,
  title={Enlightengan: Deep light enhancement without paired supervision},
  author={Jiang, Yifan and Gong, Xinyu and Liu, Ding and Cheng, Yu and Fang, Chen and Shen, Xiaohui and Yang, Jianchao and Zhou, Pan and Wang, Zhangyang},
  journal={IEEE transactions on image processing},
  volume={30},
  pages={2340--2349},
  year={2021},
  publisher={IEEE}
}

@inproceedings{zhang2021rellie,
  title={Rellie: Deep reinforcement learning for customized low-light image enhancement},
  author={Zhang, Rongkai and Guo, Lanqing and Huang, Siyu and Wen, Bihan},
  booktitle={Proceedings of the 29th ACM international conference on multimedia},
  pages={2429--2437},
  year={2021}
}

@inproceedings{liu2021retinex,
  title={Retinex-inspired unrolling with cooperative prior architecture search for low-light image enhancement},
  author={Liu, Risheng and Ma, Long and Zhang, Jiaao and Fan, Xin and Luo, Zhongxuan},
  booktitle={Proceedings of the IEEE/CVF conference on computer vision and pattern recognition},
  pages={10561--10570},
  year={2021}
}

@inproceedings{ma2022toward,
  title={Toward fast, flexible, and robust low-light image enhancement},
  author={Ma, Long and Ma, Tengyu and Liu, Risheng and Fan, Xin and Luo, Zhongxuan},
  booktitle={Proceedings of the IEEE/CVF conference on computer vision and pattern recognition},
  pages={5637--5646},
  year={2022}
}

@inproceedings{wu2022uretinex,
  title={Uretinex-net: Retinex-based deep unfolding network for low-light image enhancement},
  author={Wu, Wenhui and Weng, Jian and Zhang, Pingping and Wang, Xu and Yang, Wenhan and Jiang, Jianmin},
  booktitle={Proceedings of the IEEE/CVF conference on computer vision and pattern recognition},
  pages={5901--5910},
  year={2022}
}

@inproceedings{xu2022snr,
  title={Snr-aware low-light image enhancement},
  author={Xu, Xiaogang and Wang, Ruixing and Fu, Chi-Wing and Jia, Jiaya},
  booktitle={Proceedings of the IEEE/CVF conference on computer vision and pattern recognition},
  pages={17714--17724},
  year={2022}
}

@inproceedings{wang2022uformer,
  title={Uformer: A general u-shaped transformer for image restoration},
  author={Wang, Zhendong and Cun, Xiaodong and Bao, Jianmin and Zhou, Wengang and Liu, Jianzhuang and Li, Houqiang},
  booktitle={Proceedings of the IEEE/CVF conference on computer vision and pattern recognition},
  pages={17683--17693},
  year={2022}
}

@inproceedings{zamir2022restormer,
  title={Restormer: Efficient transformer for high-resolution image restoration},
  author={Zamir, Syed Waqas and Arora, Aditya and Khan, Salman and Hayat, Munawar and Khan, Fahad Shahbaz and Yang, Ming-Hsuan},
  booktitle={Proceedings of the IEEE/CVF conference on computer vision and pattern recognition},
  pages={5728--5739},
  year={2022}
}

@inproceedings{saharia2022palette,
  title={Palette: Image-to-image diffusion models},
  author={Saharia, Chitwan and Chan, William and Chang, Huiwen and Lee, Chris and Ho, Jonathan and Salimans, Tim and Fleet, David and Norouzi, Mohammad},
  booktitle={ACM SIGGRAPH 2022 conference proceedings},
  pages={1--10},
  year={2022}
}

@article{ozdenizci2023restoring,
  title={Restoring vision in adverse weather conditions with patch-based denoising diffusion models},
  author={{\"O}zdenizci, Ozan and Legenstein, Robert},
  journal={IEEE Transactions on Pattern Analysis and Machine Intelligence},
  volume={45},
  number={8},
  pages={10346--10357},
  year={2023},
  publisher={IEEE}
}

@inproceedings{fei2023generative,
  title={Generative diffusion prior for unified image restoration and enhancement},
  author={Fei, Ben and Lyu, Zhaoyang and Pan, Liang and Zhang, Junzhe and Yang, Weidong and Luo, Tianyue and Zhang, Bo and Dai, Bo},
  booktitle={Proceedings of the IEEE/CVF conference on computer vision and pattern recognition},
  pages={9935--9946},
  year={2023}
}

@article{jiang2023low,
  title={Low-light image enhancement with wavelet-based diffusion models},
  author={Jiang, Hai and Luo, Ao and Fan, Haoqiang and Han, Songchen and Liu, Shuaicheng},
  journal={ACM Transactions on Graphics (TOG)},
  volume={42},
  number={6},
  pages={1--14},
  year={2023},
  publisher={ACM New York, NY, USA}
}

@inproceedings{li2011image,
  title={Image enhancement based on Retinex and lightness decomposition},
  author={Li, Bo and Wang, Shuhang and Geng, Yanbing},
  booktitle={2011 18th IEEE International Conference on Image Processing},
  pages={3417--3420},
  year={2011},
  organization={IEEE}
}

@article{jobson1997properties,
  title={Properties and performance of a center/surround retinex},
  author={Jobson, Daniel J and Rahman, Zia-ur and Woodell, Glenn A},
  journal={IEEE transactions on image processing},
  volume={6},
  number={3},
  pages={451--462},
  year={1997},
  publisher={IEEE}
}

@article{abdullah2007dynamic,
  title={A dynamic histogram equalization for image contrast enhancement},
  author={Abdullah-Al-Wadud, Mohammad and Kabir, Md Hasanul and Dewan, M Ali Akber and Chae, Oksam},
  journal={IEEE transactions on consumer electronics},
  volume={53},
  number={2},
  pages={593--600},
  year={2007},
  publisher={IEEE}
}

@article{kingma2021variational,
  title={Variational diffusion models},
  author={Kingma, Diederik and Salimans, Tim and Poole, Ben and Ho, Jonathan},
  journal={Advances in neural information processing systems},
  volume={34},
  pages={21696--21707},
  year={2021}
}

@inproceedings{wang2024zero,
  title={Zero-reference low-light enhancement via physical quadruple priors},
  author={Wang, Wenjing and Yang, Huan and Fu, Jianlong and Liu, Jiaying},
  booktitle={Proceedings of the IEEE/CVF conference on computer vision and pattern recognition},
  pages={26057--26066},
  year={2024}
}

@inproceedings{jiang2024lightendiffusion,
  title={Lightendiffusion: Unsupervised low-light image enhancement with latent-retinex diffusion models},
  author={Jiang, Hai and Luo, Ao and Liu, Xiaohong and Han, Songchen and Liu, Shuaicheng},
  booktitle={European Conference on Computer Vision},
  pages={161--179},
  year={2024},
  organization={Springer}
}

@inproceedings{yi2023diff,
  title={Diff-retinex: Rethinking low-light image enhancement with a generative diffusion model},
  author={Yi, Xunpeng and Xu, Han and Zhang, Hao and Tang, Linfeng and Ma, Jiayi},
  booktitle={Proceedings of the IEEE/CVF international conference on computer vision},
  pages={12302--12311},
  year={2023}
}

\end{document}